\documentclass{article}

\PassOptionsToPackage{numbers, compress}{natbib}
\usepackage[main, final]{neurips_2026}

\usepackage[utf8]{inputenc} 
\usepackage[T1]{fontenc}    
\usepackage{hyperref}       
\usepackage{url}            
\usepackage{booktabs}       
\usepackage{amsfonts}       
\usepackage{nicefrac}       
\usepackage{microtype}      
\usepackage{xcolor}         

\hypersetup{hidelinks, pdftitle={MapPFN: Learning Causal Perturbation Maps in Context}, pdfauthor={Marvin Sextro, Weronika Kłos, Gabriel Dernbach}}       
\usepackage{graphicx}
\usepackage{subcaption}
\usepackage{amsmath}
\usepackage{amssymb}
\usepackage{mathtools}
\usepackage{amsthm}
\usepackage{xspace}
\usepackage{bm}
\usepackage{titletoc}        
\usepackage{multirow}
\usepackage{wrapfig}         
\usepackage{float}           
\usepackage{pifont}          
\usepackage[textsize=tiny]{todonotes}
\usepackage{algorithm}
\usepackage{algorithmic}
\usepackage[capitalize,noabbrev]{cleveref}   

\theoremstyle{plain}

\theoremstyle{definition}

\theoremstyle{remark}

\newcommand{\yobs}{\mathbf{y}^\text{obs}}
\newcommand{\yint}{\mathbf{y}^\text{int}}
\newcommand{\Yobs}{\mathbf{Y}^\text{obs}}
\newcommand{\Yint}{\mathbf{Y}^\text{int}}
\newsavebox\CBox
\def\textBF#1{\sbox\CBox{#1}\resizebox{\wd\CBox}{\ht\CBox}{\textbf{#1}}}
\newcommand{\yes}{{\color{green!60!black}\ding{51}}}
\newcommand{\no}{{\color{red!70!black}\ding{55}}}

\title{MapPFN: Learning Causal\\Perturbation Maps in Context}

\author{%
  Marvin Sextro\textsuperscript{\normalfont 1,2,3} \quad
  Weronika K\l{}os\textsuperscript{\normalfont 1,2} \quad
  Gabriel Dernbach\textsuperscript{\normalfont 1,4,2,3} \\[0.8ex]
  \textsuperscript{1}Machine Learning Group, Technische Universität Berlin, Berlin, Germany \\
  \textsuperscript{2}BIFOLD -- Berlin Institute for the Foundations of Learning and Data, Berlin, Germany \\
  \textsuperscript{3}Aignostics, Berlin, Germany \\
  \textsuperscript{4}Institute of Pathology, Charité -- Universitätsmedizin Berlin, Berlin, Germany \\
  \texttt{\{m.kleine.sextro, w.klos\}@tu-berlin.de, gabriel.dernbach@gmail.com}
}

\begin{document}

\maketitle

\begin{abstract}

Planning effective interventions in biological systems requires treatment-effect models that adapt to unseen biological contexts by identifying their specific underlying mechanisms. Yet single-cell perturbation datasets span only a handful of biological contexts, and existing methods cannot leverage new interventional evidence at inference time to adapt beyond their training data. To meta-learn a perturbation effect estimator, we present MapPFN, a prior-data fitted network (PFN) pre-trained on a synthetic biological prior with causal interventions, decoupling pre-training from limited wet-lab data. Unlike existing methods, MapPFN uses in-context learning to map a sequence of experiments to a post-perturbation distribution, enabling a single pre-trained model to adapt to new datasets and gene sets. Our results show that combining synthetic pre-training with fine-tuning on real single-cell data improves perturbation predictions across biological contexts. The code, model and data are available at \url{https://marvinsxtr.github.io/MapPFN}.

\end{abstract}
\begin{figure*}[t]
  \vskip 0.2in
  \begin{center}
    \centerline{\includegraphics[width=0.96\textwidth]{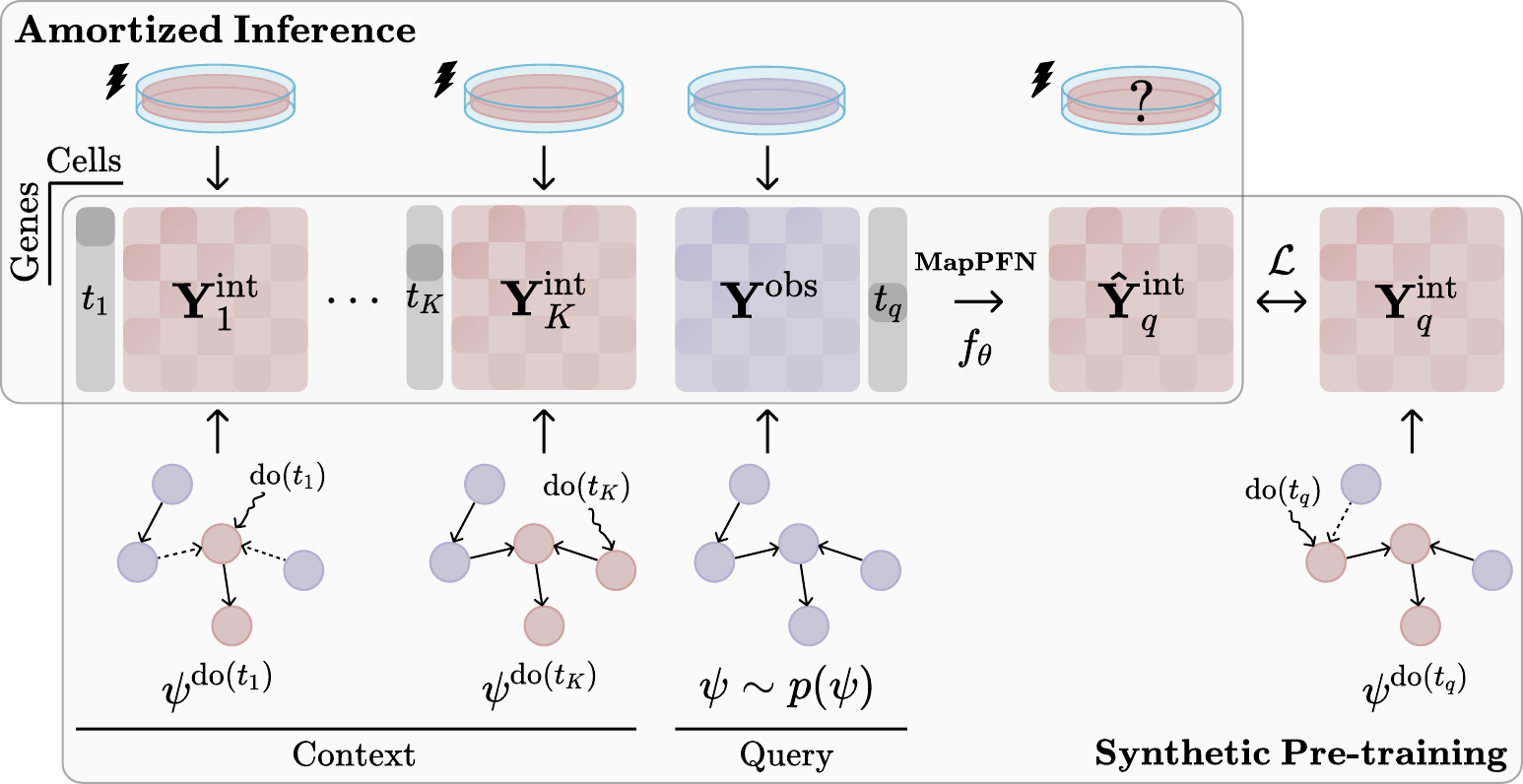}}
    \caption{\textbf{MapPFN overview.} MapPFN uses in-context learning (ICL) to predict perturbation effects in unseen biological contexts. During pre-training, we draw structural causal models (SCMs) or synthetic gene regulatory networks (GRNs) $\psi$ to generate samples from the observational distribution $\Yobs$ and a context set of interventional distributions $\mathcal{C}=\{(t_k, \Yint_k)\}_{k=1}^K$, where $t_k$ denotes a perturbation (do-intervention). Given $\smash{\Yobs}$ and the context set $\mathcal{C}$, MapPFN predicts post-perturbation distributions $\Yint_q$ arising from unseen interventions $t_q$. During pre-training, MapPFN meta-learns how to map between pre- and post-perturbation distributions across many causal structures $\psi$ by minimizing $\mathcal{L}(\hat{\mathbf{Y}}_q^{\text{int}}, \Yint_q)$. At inference time, MapPFN predicts cell-level post-perturbation distributions $\Yint_q \in \mathbb{R}^{\text{cells} \times \text{genes}}$ in one step through amortized inference, without requiring knowledge of the underlying causal structure $\psi$.}
    \label{fig:mappfn_overview}
  \end{center}
  \vskip -0.2in
\end{figure*}

\section{Introduction}

To gain a mechanistic understanding of the behavior of cell populations, single-cell perturbation data has long been the experimental gold standard to identify the causal dependencies that form underlying gene regulatory networks (GRNs) \citep{sachs_causal_2005}. Genetic CRISPR knockout perturbations \citep{jinek_programmable_2012} measured in single cells using Perturb-Seq \citep{dixit_perturb-seq_2016} allow us to measure the outcome of targeted interventions in controlled biological contexts, such as cell lines and their cell states \citep{frangieh_multimodal_2021, papalexi_characterizing_2021}. However, mapping the whole space of possible biological contexts and perturbations through experiments alone is infeasible.

Virtual cell models aim to reduce the costs of drug target discovery by predicting how cells respond to small molecules or gene knockouts \citep{bunne_how_2024, roohani_virtual_2025}, enabling high-throughput evaluation of hypotheses prior to time-consuming validation in the wet lab. In practice, such models remain constrained by data scarcity, as even the largest perturbation dataset to date covers only 50 cell lines \citep{zhang_tahoe-100m_2026}.

Because sequencing destroys individual cells, perturbation prediction becomes a problem of mapping between unpaired distributions, making optimal transport (OT) a natural approach. These methods learn a transport map between the pre- and post-perturbation cell distributions, conditioned on a treatment or covariates \citep{bunne_learning_2023, dong_causal_2023}. Lifting the strict assumptions of OT-based methods, recent approaches use generative models to predict the post-perturbation distribution conditioned on covariates \citep{lotfollahi_predicting_2023, klein_cellflow_2025} or a learned representation of the initial observational distribution \citep{atanackovic_meta_2025, adduri_predicting_2025}. Yet they lack test-time adaptation from a sequence of interventional distributions, constraining generalization to the biological contexts seen during training.

In this work, we propose to meta-learn perturbation maps from a multi-experiment input of observational and interventional distributions, enabling a diffusion transformer to infer perturbation effects via in-context learning. Building on the recent success of prior-data fitted networks (PFNs) \citep{muller_transformers_2022} in tabular prediction \citep{hollmann_accurate_2025, hollmann_tabpfn_2023, qu_tabicl_2025} and causal inference \citep{robertson_do-pfn_2025, balazadeh_causalpfn_2025, ma_foundation_2026}, we introduce the first PFN for perturbation prediction, pre-trained on data generated from a synthetic biological prior. In contrast to standard PFN training, our task requires predicting a distribution of vectors, for which we adopt the multimodal diffusion transformer (MMDiT) \citep{esser_scaling_2024} architecture. We show that conditioning on pre- and multiple post-perturbation distributions improves performance over models that only condition on a pre-perturbation distribution with a query treatment identifier. With MapPFN, we demonstrate that combining pre-training on a synthetic biological prior with fine-tuning on real single-cell data improves perturbation predictions across biological contexts~\citep{frangieh_multimodal_2021,papalexi_characterizing_2021}.

\paragraph{Our Contributions}

\begin{enumerate}
    \item We frame perturbation prediction as a distribution mapping with test-time interventional context, enabling a single pre-trained model to adapt to new biological contexts and gene sets via in-context learning.
    \item We introduce MapPFN, the first prior-data fitted network (PFN) for perturbation prediction. Pre-trained on a synthetic biological prior of \emph{in silico} gene knockouts, MapPFN meta-learns perturbation maps across diverse causal structures and is not limited by the availability of experimental perturbation data.
    \item In a controlled synthetic benchmark of structural causal models (SCMs) with known mechanisms, MapPFN successfully meta-learns perturbation prediction. Our ablations show that interventional context and a counterfactual prior each independently improve predictions.
    \item Evaluating MapPFN on biologically distinct perturbation datasets, we show that combining synthetic pre-training with fine-tuning on real single-cell data improves perturbation predictions across biological contexts.
\end{enumerate}

\section{Problem Statement}

We consider the problem of learning how biological systems behave under interventions. In the case of single-cell perturbations, we are given a set of $N$ gene expression profiles $\yobs \in \mathbb{R}^d$ measured in a specific biological context and a treatment $t\in\mathcal{T}$ in the form of an intervention on a single gene, resulting in $M$ post-treatment gene expression profiles $\yint \in \mathbb{R}^d$. The resulting dataset takes the form $\{(\Yobs_\ell, t_\ell, \Yint_\ell)\}_{\ell=1}^L$, where $\Yobs \in \mathbb{R}^{N\times d}$, $\Yint \in \mathbb{R}^{M\times d}$ and $L$ is the number of pairs of biological contexts and treatments. Importantly, there is no direct correspondence between any two pre- and post-treatment cells, rendering this a problem of learning a map between distributions $p(\yobs)$ and $p(\yint)$.

The same intervention can produce different perturbation responses depending on the biological context and its underlying causal mechanisms. We therefore condition on observational samples $\Yobs$ and an interventional context $\mathcal{C}=\{(t_k, \Yint_k)\}_{k=1}^K$ for a subset of treatment conditions $t_k \in \mathcal{T}_\mathcal{C} \subset \mathcal{T}$ for a given biological context, and aim to predict the outcome distribution of an unseen query perturbation $t_q \in \mathcal{T} \setminus \mathcal{T}_\mathcal{C}$:
\begin{equation}
    p(\yint_q \mid \text{do}(t_q), \Yobs, \mathcal{C})
\end{equation}

\section{Background and Related Work}

\paragraph{Perturbation Prediction} Existing methods differ in their generalization target and conditioning capabilities. Approaches like CPA \citep{lotfollahi_predicting_2023}, CellOT \citep{bunne_learning_2023} and CellFlow \citep{klein_cellflow_2025} condition on covariates and aim to generalize across biological contexts. Meta Flow Matching \citep{atanackovic_meta_2025} and STATE \citep{adduri_predicting_2025} additionally condition on the observational distribution. Methods targeting unseen perturbations instead make assumptions about the causal structure, either through explicit modeling \citep{schneider_generative_2025} or by incorporating known GRNs \citep{roohani_predicting_2024}. Single-cell foundation models \citep{theodoris_transfer_2023, cui_scgpt_2024, hao_large-scale_2024} perform perturbation effect analysis on individual cells rather than generating post-perturbation distributions. Our work targets generalization to new biological contexts and gene sets, requiring no knowledge of the underlying causal structure.

\paragraph{Amortized and In-Context Learning} Rather than optimizing per task, amortized methods learn to perform inference in a single forward pass conditioned on a task context. This context can take the form of the whole dataset for causal structure learning \citep{lorch_amortized_2022, ke_learning_2023, dhir_a_2025} or an input distribution for OT \citep{amos_meta_2023, klein_genot_2024} or generative modeling \citep{atanackovic_meta_2025}. Exemplified by large language models \citep{brown_language_2020}, in-context learning (ICL) achieves amortization by conditioning on example tasks in the input sequence. Recent evidence shows that next-token prediction alone can induce causal discovery and counterfactual reasoning in transformers \citep{butkus_causal_2025}. Concurrent to our work, \citet{dong_stack_2026} apply ICL to single-cell perturbation prediction. Unlike our approach, they limit the interventional context set to a single experiment and do not use a synthetic prior for pre-training.

\paragraph{Prior-data Fitted Networks} Prior-data fitted networks (PFNs) are pre-trained on synthetic datasets to perform Bayesian inference in context \citep{muller_transformers_2022}. In a classical supervised machine learning setting with a dataset $\mathcal{D} = \{(\mathbf{x}_i,y_i)\}_{i=1}^N$, Bayesian inference assumes a prior $p(\psi)$ representing a space of hypotheses (e.g.\ structural causal models) that could have generated the data. The aim of PFNs is to approximate the posterior predictive distribution (PPD) $p(y \mid \mathbf{x}, \mathcal{D})$. Given a complete training dataset $\mathcal{D} = \{(\mathbf{x}_i,y_i)\}_{i=1}^N$ and an unlabeled query $\mathbf{x}_q$ from the test set, a PFN directly outputs the predicted label $y_q$. Since the learning process happens in the context of a transformer within a single forward pass, this process is regarded as in-context learning or amortized Bayesian inference. Training PFNs involves sampling a large number of hypotheses $\psi \sim p(\psi)$ and generating synthetic datasets $\mathcal{D} \sim p(\mathcal{D} \mid \psi)$ in an outer loop to meta-learn how to make predictions in context. We refer to \citet{muller_transformers_2022} and \citet{hollmann_tabpfn_2023} for further details.

PFNs have recently surpassed classical methods in tabular prediction benchmarks \citep{hollmann_accurate_2025} and have been applied to other problems, including causal inference \citep{balazadeh_causalpfn_2025, robertson_do-pfn_2025, ma_foundation_2026}, full Bayesian inference \citep{reuter_can_2025} and optimization \citep{muller_pfns4bo_2023}. Yet, contrary to our approach, existing PFNs for causal inference only predict univariate outcomes for individual samples rather than population-level distributions, rendering them incapable of handling perturbation data. In addition, they condition predictions only on observational data and a query treatment, but not on interventional data.

\section{Priors for Perturbation Prediction}

We use two priors corresponding to different evaluation settings. To evaluate MapPFN in a controlled environment where the causal mechanism is known, we sample structural causal models (SCMs). For real-data inference where the causal mechanism is unknown, we sample gene regulatory networks (GRNs) from a synthetic biological prior with nonlinear Hill functions. Additional details on the priors are provided in Section~\ref{sec:priors}.

\paragraph{Structural Causal Models} A structural causal model (SCM) $\psi$ \citep{pearl_causality_2009} defines a generative model through a directed acyclic graph (DAG) $\mathcal{G}_\psi$ over variables $\{z_1, z_2, ..., z_d\}$, together with structural assignment $z_k = f_k(z_{\mathrm{PA}(k)}, \epsilon_k)$ for each node $z_k$, where $z_{\mathrm{PA}(k)}$ denotes the parents of $z_k$ in $\mathcal{G}_\psi$, $f_k$ is a deterministic function, and $\epsilon_k$ is an exogenous noise variable. Following the rules of do-calculus \citep{pearl_causality_2009}, a hard intervention $\text{do}(t)$ on node $z_k$ removes its incoming edges and assigns $z_k := t$, yielding $\psi^{\text{do}(t)}$. Linear additive noise models (ANMs) are a class of SCMs with linear functional relationships $f_k$ and additive noise. In this case, the model is fully determined by a sparse weighted adjacency matrix $\mathbf{W}\in \mathbb{R}^{d\times d}$, where $w_{kj} \neq 0$ only if $j \in \mathrm{PA}(k)$. Given a noise vector $\bm{\epsilon} \sim \mathcal{N}(0, \mathbf{I})$, we can sample from linear ANMs by solving the linear system $\mathbf{z} = (\mathbf{I} - \mathbf{W})^{-1}\bm{\epsilon}$~\citep{pearl_causality_2009}.

\paragraph{Synthetic Biological Prior} Since cells from the same cell line are genetic clones drawn from a single regulatory mechanism, experimental perturbation screens provide many cells but few distinct causal structures. In practice, even the largest dataset to date contains 100 million cells but only 50 cell lines \citep{zhang_tahoe-100m_2026}. We decouple MapPFN from this bottleneck by pre-training on synthetic data generated from a biological prior based on established components validated against single-cell screens \citep{dibaeinia_sergio_2020, aguirre_gene_2025}. Specifically, we sample diverse gene regulatory networks (GRN) with realistic sparsity and modular structure, from which we simulate observational and interventional gene expression dynamics.

\section{Meta-Learning Perturbation Prediction with MapPFN}

Below we describe MapPFN using SCMs as the running example. The same procedure applies to the synthetic biological prior, simulating nonlinear gene expression dynamics instead of linear SCMs.

\paragraph{Modeling Assumptions} We assume the observations $\Yobs$ are generated by a latent SCM $\psi$. We consider single-node hard interventions $t \in \mathcal{T}$, where each treatment corresponds to a gene knockout modeled as $\text{do}(t)$ on the underlying causal structure. We assume $\Yint$ to stem from the intervened-upon SCM $\psi^{\text{do}(t)}$ and that all variables of the latent SCM are observed.

Given observational samples $\Yobs$ and a set of interventional experiments $\mathcal{C} = \{(t_k, \Yint_k)\}_{k=1}^K$ for $\psi$, we aim to directly predict the post-perturbation distribution of an unseen query treatment $t_q \in \mathcal{T} \setminus \mathcal{T}_\mathcal{C}$. Based on our assumptions, the posterior predictive distribution takes the form
\begin{equation}
    p(\mathbf{y}_q^\text{int} \mid \text{do}(t_q), \Yobs, \mathcal{C}) = \int p(\mathbf{y}_q^\text{int} \mid \text{do}(t_q), \Yobs, \psi) \, p(\psi \mid \Yobs,  \mathcal{C}) \, d\psi
\end{equation}
MapPFN approximates this distribution by amortizing inference over diverse causal structures $\psi$ sampled during synthetic pre-training. We refer to \citet{robertson_do-pfn_2025} for a theoretical discussion of the sources of uncertainty in this formulation.

In contrast to existing methods, MapPFN does not require a data split across multiple biological contexts, adapting to the context at hand from a set of observational and interventional distributions via in-context learning. Additionally, existing models must be retrained on each new gene set, whereas MapPFN supports an arbitrary selection of genes by pre-training on \emph{in silico} knockouts.

\begin{wrapfigure}{R}{0.51\textwidth}
\hrule height 0.8pt
\vspace{2pt}
\captionsetup{singlelinecheck=false, justification=raggedright, aboveskip=0pt, belowskip=0pt}
\captionof{algorithm}{\textbf{MapPFN Pre-training}}\label{alg:mappfn_pretraining}
\vspace{2pt}
\hrule height 0.4pt
\vspace{2pt}
\small
\begin{algorithmic}
  \STATE {\bfseries Input:} prior $p(\psi)$, treatments $\mathcal{T}$, context size $K$
  \FOR{$i=1, 2, \ldots, N$}
  \STATE Draw SCM $\psi \sim p(\psi)$
  \STATE Draw observational samples $\Yobs \sim p(\yobs \mid \psi)$
  \STATE Draw context treatments $\mathcal{T}_\mathcal{C} \subset \mathcal{T}$ with $|\mathcal{T}_\mathcal{C}| = K$
  \FOR{$k=1, \ldots, K$}
  \STATE Draw $\Yint_k \sim p(\yint \mid \text{do}(t_k), \psi)$
  \ENDFOR
  \STATE Set context $\mathcal{C} \leftarrow \{(t_k, \Yint_k)\}_{k=1}^K$
  \STATE Draw query treatment $t_q \sim \mathcal{T} \setminus \mathcal{T}_\mathcal{C}$
  \STATE Draw target $\Yint_q \sim p(\yint \mid \text{do}(t_q), \psi)$
  \STATE Draw time $\tau \sim \text{LogitNormal}(0,1)$, $\mathbf{Y}_0 \sim \mathcal{N}(0, \mathbf{I})$
  \STATE Compute $\mathcal{L}_{\text{CFM}}(\theta; \mathbf{Y}_0, \tau, \Yint_q, t_q, \Yobs, \mathcal{C})$
  \STATE Update $\theta \leftarrow \theta - \alpha \nabla \mathcal{L}_{\text{CFM}}(\theta)$
  \ENDFOR
\end{algorithmic}
\vspace{2pt}
\hrule height 0.8pt
\vspace{2pt}
{\footnotesize\textit{Note:} $\mathbf{Y} \sim p(\mathbf{y} \mid \cdot, \psi)$ implies first sampling noise $\mathbf{N} \in \mathbb{R}^{n\times d}$ and stacking $n$ i.i.d.\ samples.}
\vspace{-2em}
\end{wrapfigure}

\paragraph{Pre-training Process} During each pre-training step, we first sample an SCM $\psi \sim p(\psi)$ from the prior. By propagating noise $\mathbf{N} = [\bm{\epsilon}_1, ..., \bm{\epsilon}_n]^\top, \bm{\epsilon}_i \sim \mathcal{N}(0, \mathbf{I})$ through the SCM, we obtain the observational distribution $\Yobs$. Subsequently, we build the context $\mathcal{C} = \{(t_k, \Yint_k)\}_{k=1}^K$ by sampling SCMs $\psi^{\text{do}(t_k)}$ for a subset of treatments $t_k \in \mathcal{T}_\mathcal{C} \subset \mathcal{T}$. For each intervention in this set, we generate post-perturbation distributions $\Yint_k$ by drawing new noise $\mathbf{N}_k$. Finally, our prediction target is the post-perturbation distribution $\Yint_q$ arising from an unseen query treatment $t_q \in \mathcal{T} \setminus \mathcal{T}_\mathcal{C}$. \autoref{fig:mappfn_overview} provides an overview of MapPFN pre-training and inference. The full pre-training process is outlined in Algorithm~\ref{alg:mappfn_pretraining}.

\paragraph{Identifiability} Perturbation prediction depends on identifiability, i.e.\ the extent to which the causal graph $\mathcal{G}_\psi$ can be inferred from data, even if it is not explicitly recovered. Interventional data can fully identify the causal graph given sufficient interventions \citep{eberhardt_n-1_2006}. Conditioning on an interventional context $\mathcal{C}$ reduces the Markov equivalence class $[\mathcal{G}_\psi]$, as each intervention constrains the set of causal structures consistent with the data \citep{hauser_characterization_2012}. This provides MapPFN with a theoretical advantage over existing causal PFNs and perturbation models that learn from observational data alone, assuming the true causal graph lies within the support of the prior distribution $p(\psi)$.

\paragraph{Model} We adopt the multimodal diffusion transformer (MMDiT) \citep{esser_scaling_2024} architecture with minor modifications. Treating cells as tokens, we process input noise, expression profiles and one-hot encoded treatments as three modality streams with separate parameters. Cross-modal interactions are enabled via joint attention. We compare MMDiT against a diffusion transformer (DiT) \citep{peebles_scalable_2023} with shared weights in Appendix~\ref{sec:architecture_ablation}.

Because the inputs are unordered sets of cells, we remove sinusoidal positional encodings and rely on the permutation invariance of attention. Instead, we introduce learnable embeddings to differentiate modalities, query versus context, and observational versus interventional data. We train MapPFN using a conditional flow matching objective \citep{lipman_flow_2023}, which learns a velocity field that transports noise to the predicted post-perturbation distribution.
\begin{equation}
    \mathcal{L}_\text{CFM}(\theta) = \mathbb{E}_{\tau, \mathbf{Y}_0, \Yint_q} \left\| v^\theta_\tau(\mathbf{Y}_\tau \mid t_q, \Yobs, \mathcal{C}) - (\Yint_q - \mathbf{Y}_0) \right\|^2_\text{F}
\end{equation}
where $v^\theta_\tau$ is the learned velocity, $\mathbf{Y}_\tau = (1-\tau)\mathbf{Y}_0 + \tau \Yint_q$ is the interpolated sample at time $\tau$, and $\theta$ denotes the model parameters. Please refer to \autoref{sec:model} for details on the model architecture.

\section{Experimental Setup}

We evaluate MapPFN in a controlled environment of known linear SCMs and on real-world single-cell perturbation datasets. For linear SCMs, we train and evaluate all methods including MapPFN on data from the same synthetic prior. For the single-cell experiments, our evaluation setting for all methods follows the Virtual Cell Challenge \citep{roohani_virtual_2025}, where adaptation is based on a limited number of interventional experiments in the holdout context. MapPFN is pre-trained on the synthetic biological prior and optionally fine-tuned on real perturbation data. In contrast, baselines are trained from scratch on real single-cell data, including the perturbations MapPFN receives as interventional context. Additional details on the setup are provided in Appendix~\ref{sec:experimental_details}.

\subsection{Priors} \label{sec:priors}

\paragraph{Structural Causal Models} We generate synthetic data from linear structural causal models (SCM) with additive Gaussian noise \citep{pearl_causality_2009}. We sample directed acyclic graphs (DAGs) from an Erd\H{o}s--R\'enyi distribution \citep{erdos_evolution_1960} with $d=20$ nodes and an edge probability of $p=0.5$. Additional details on the linear SCM data are provided in Appendix~\ref{sec:synthetic_scm_data}.

\paragraph{Synthetic Biological Prior} We first generate directed graphs from a scale-free distribution using the preferential attachment algorithm \citep{aguirre_gene_2025}, allowing us to generate networks with similar properties to real GRNs in terms of modularity, sparsity and degree distributions (see Appendix~\ref{sec:synthetic_rna_data}).

Given a sampled regulatory network, we simulate single-cell expression profiles using SERGIO \citep{dibaeinia_sergio_2020}, which models gene expression as the steady state of a system of stochastic differential equations (SDEs). Regulatory interactions are parameterized by Hill functions \citep{gesztelyi_hill_2012}, capturing nonlinear and saturation effects. Genetic perturbations are performed \emph{in silico} by removing the perturbed gene from the regulatory network and re-simulating the system. To obtain gene expression counts, we apply the technical noise model of SERGIO for 10x Chromium single-cell RNA sequencing. Additional details on the synthetic biological prior and its hyperparameters are provided in Appendix~\ref{sec:synthetic_rna_data}.

\subsection{Single-cell Perturbation Datasets} \label{sec:datasets}

We evaluate MapPFN on two biologically distinct single-cell perturbation datasets. The first \citep{frangieh_multimodal_2021} consists of approximately 218,000 cells from a CRISPR knockout screen of 248 genes in melanoma cells across three treatment-induced cell states. The second \citep{papalexi_characterizing_2021} consists of approximately 20,000 cells from a CRISPR perturbation screen of 25 genes in a leukemia cell line. Following \citet{schneider_generative_2025}, we focus our analysis on 50 genes for both datasets. Additional details on the perturbation datasets are provided in Appendix~\ref{sec:single_cell_data}.

\begin{wraptable}{R}{0.475\textwidth}
\vspace{-1.4em}
\caption{\textbf{Overview of perturbation models by conditioning capability.} Columns indicate whether a method conditions on covariates, observational populations, or interventional populations. While all baselines require interventional data as a supervision target, MapPFN uniquely conditions on interventional data via in-context learning.}
\label{tab:conditioning_overview}
\centering
\resizebox{\linewidth}{!}{%
\begin{tabular}{lccc}
\toprule
Methods & Covariates & Observational & Interventional \\
\midrule
CPA, CondOT, CellFlow & \yes & \no & \no \\
MFM, STATE & \yes & \yes & \no \\
\textbf{MapPFN (ours)} & \yes & \yes & \yes \\
\bottomrule
\end{tabular}%
}
\end{wraptable}

\subsection{Baselines} We compare our method against CPA \citep{lotfollahi_predicting_2023}, Conditional Optimal Transport (CondOT) \citep{bunne_supervised_2022}, Meta Flow Matching (MFM) \citep{atanackovic_meta_2025}, CellFlow \citep{klein_cellflow_2025} and STATE \citep{adduri_predicting_2025}. While these baselines condition on covariates or observational data and use interventional data as a supervision target, MapPFN conditions on interventional data via in-context learning (see \autoref{tab:conditioning_overview}). As lower and upper bounds, we report two reference baselines following \citet{bunne_learning_2023}: an identity baseline that predicts the observational distribution $\hat{\mathbf{y}}^\text{int} \sim p(\yobs)$, and an oracle baseline that uses the observed distribution $\hat{\mathbf{y}}^\text{int} \sim p(\yint)$. Following \citet{ahlmann-eltze_deep-learning-based_2025}, we report the training perturbation mean and a linear baseline over PCA-derived gene and perturbation embeddings. Additional details on the baselines are provided in Appendix~\ref{sec:baselines}.

\subsection{Metrics} We evaluate model performance by comparing the predicted post-perturbation distribution $\hat{\mathbf{Y}}^\text{int}$ to the ground-truth distribution $\Yint$ in terms of distributional similarity, moment-level accuracy, perturbation discrimination and differentially expressed gene (DEG) recovery. Distributional similarity is quantified using the entropy-regularized Wasserstein distance ($\text{W}_2$) \citep{cuturi_sinkhorn_2013} and the maximum mean discrepancy (MMD) \citep{gretton_kernel_2012}. Moment-level accuracy is measured by the root mean squared error (RMSE) between the predicted and ground-truth distribution means. To assess whether predictions are distinguishable across perturbations, we report the ranking-based perturbation discrimination score (PDS) \citep{wu_perturbench_2025}. Identifying which genes are differentially expressed is critical for understanding treatment mechanisms and planning interventions. We therefore evaluate DEG recovery using the area under the precision-recall curve (AUPRC) \citep{zhu_auprc_2025}, comparing DEGs from the predicted post-perturbation distribution with those observed in the ground-truth data. Additional details on the metrics are provided in Appendix~\ref{sec:metrics}.

\paragraph{Magnitude Ratio} Causal effects can occur on different scales across biological contexts, making absolute distributional distances difficult to interpret. In particular, a small distance does not imply a weak causal effect, nor does a large distance imply a strong one. To normalize for effect scale, we introduce the \emph{magnitude ratio} (MR), which measures how much of the true intervention effect is recovered by the prediction. Let $d$ denote a distributional distance (e.g.\ Wasserstein distance). The magnitude ratio is defined as
\begin{equation}
    \text{MR}(\Yobs, \Yint, \hat{\mathbf{Y}}^\text{int}) = \frac{d(\Yobs, \hat{\mathbf{Y}}^\text{int})}{d(\Yobs, \Yint)}
\end{equation}
A perfect prediction corresponds to a magnitude ratio of $1.0$ and an identity collapse ($\hat{\mathbf{Y}}^\text{int} = \Yobs$) results in a magnitude ratio of 0.0. The magnitude ratio is invariant to the absolute effect scale and diagnoses effect size collapse but is insensitive to directionality, which the remaining metrics capture. Throughout, we report it using the Wasserstein distance.

\section{Results}

\begin{figure}[t]
\vskip 0.1in
\begin{minipage}[t]{0.48\textwidth}
  \centering
  \includegraphics[width=\linewidth]{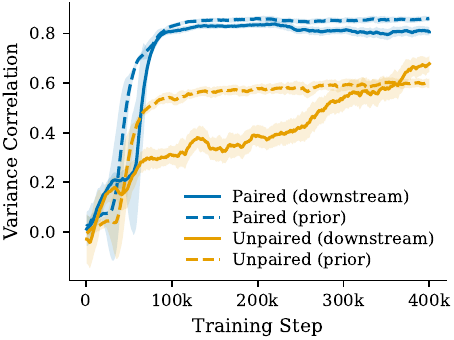}
  \caption{\textbf{Counterfactual paired prior improves downstream performance.} The paired prior converges faster and to higher performance than the unpaired prior. Variance correlation measures the Pearson correlation between feature variances of predicted and ground-truth samples. Shaded areas indicate a rolling standard deviation over 50 validation steps.}
  \label{fig:prior_convergence}
\end{minipage}\hfill
\begin{minipage}[t]{0.48\textwidth}
  \centering
  \includegraphics[width=\linewidth]{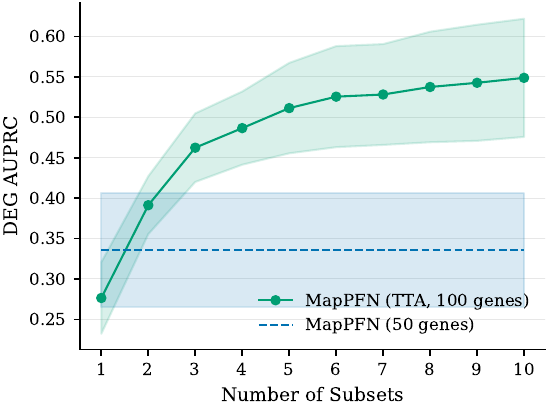}
  \caption{\textbf{MapPFN scales to larger gene sets via test-time augmentation.} MapPFN (TTA) aggregates predictions over ten random overlapping subsets of 50 genes drawn from 100 genes, improving recovery of differentially expressed genes as measured by AUPRC. Shaded areas indicate standard deviation over five data splits, each averaged over ten resampling seeds.}
  \label{fig:tta}
\end{minipage}
\end{figure}

We report benchmarking results for linear SCMs in \autoref{tab:scm_unpaired} and for single-cell datasets in \autoref{tab:singlecell_benchmark}. We ablate the interventional context and counterfactual paired prior in \autoref{tab:frangieh_ablation} and \autoref{fig:prior_convergence}, and demonstrate scaling to larger gene sets via test-time augmentation in \autoref{fig:tta}.

\paragraph{MapPFN successfully meta-learns perturbation prediction in a controlled environment} \autoref{tab:scm_unpaired} compares MapPFN against CondOT and MFM within a prior of linear SCMs. MapPFN achieves the best performance across metrics, only tied with MFM on Wasserstein distance. The magnitude ratio reveals identity collapse as a common failure mode in baselines. CondOT and MFM yield magnitude ratios around 0.1, suggesting little deviation from the observational distribution, while MapPFN is the only method with a magnitude ratio close to one. We attribute this to both baselines either initializing at the observational distribution or initializing the model weights as an identity map. These results motivate evaluation on real single-cell data where the underlying causal structure is unknown.

\begin{table*}[h]
\centering
\caption{\textbf{Evaluation of MapPFN within a prior of linear SCMs.} Mean $\pm$ std over three training seeds. Bold indicates results within one standard deviation of the best. In this controlled setting, MapPFN successfully meta-learns perturbation prediction.}
\label{tab:scm_unpaired}
\resizebox{0.7\textwidth}{!}{%
\begin{tabular}{lccccc}
\toprule
Method & $\text{W}_2$ $\downarrow$ & MMD ($\times 10^{-3}$) $\downarrow$ & RMSE $\downarrow$ & PDS $\downarrow$ & MR $\rightarrow 1$ \\
\midrule
CondOT \citep{bunne_supervised_2022} & $13.85 \pm 0.12$ & $5.14 \pm 0.01$ & $0.15 \pm 0.00$ & $0.11 \pm 0.02$ & $0.09 \pm 0.01$ \\
MFM \citep{atanackovic_meta_2025} & $\textBF{13.73} \pm 0.16$ & $4.81 \pm 0.19$ & $0.15 \pm 0.01$ & $0.09 \pm 0.03$ & $0.12 \pm 0.00$ \\
MapPFN (ours) & $\textBF{13.69} \pm 0.05$ & $\textBF{4.28} \pm 0.06$ & $\textBF{0.14} \pm 0.00$ & $\textBF{0.01} \pm 0.01$ & $\textBF{0.99} \pm 0.02$ \\
\midrule
Identity & $17.61 \pm 0.14$ & $12.98 \pm 0.35$ & $0.28 \pm 0.01$ & $0.49 \pm 0.01$ & $0.00 \pm 0.00$ \\
Observed & $9.82 \pm 0.08$ & $3.66 \pm 0.06$ & $0.07 \pm 0.00$ & $0.00 \pm 0.00$ & $1.00 \pm 0.00$ \\
\bottomrule
\end{tabular}}
\end{table*}

\paragraph{Synthetic pre-training enables adaptation to new datasets and gene sets} \autoref{tab:singlecell_benchmark} compares a single MapPFN pre-trained on synthetic data, adapting to each dataset via in-context learning, against baselines trained from scratch on two biologically distinct single-cell perturbation datasets. On the melanoma dataset, MapPFN recovers differentially expressed genes and preserves perturbation effect magnitudes, outperforming CPA, CellFlow, Mean and Linear on AUPRC and PDS and achieving the best MR, while MMD and RMSE remain above the Identity baseline. On the leukemia dataset, distributional metrics degrade and MR indicates an overshoot of the ground-truth effect, which we attribute to the dataset containing only 23 perturbation targets, with the remainder being downstream marker genes. In contrast, both the synthetic prior and the melanoma dataset consist of genes within a shared regulatory mechanism (\autoref{fig:prior_coverage} in \autoref{sec:additional_results}).

\paragraph{Fine-tuning improves perturbation prediction performance across biological contexts} Fine-tuned MapPFN matches or outperforms all baselines on all metrics except $\text{W}_2$ across both datasets, yielding improvements over pre-trained MapPFN across metrics and datasets (\autoref{tab:singlecell_benchmark}). To assess the contribution of synthetic pre-training, we compare fine-tuned MapPFN against a randomly initialized variant trained directly on real single-cell perturbation data, without synthetic pre-training (\autoref{tab:frangieh_ablation}). Random initialization performs worse on all metrics, suggesting that meta-learning perturbation prediction on diverse synthetic causal mechanisms improves performance.

\begin{table*}
\centering
\caption{\textbf{Comparison of MapPFN against baselines across two single-cell perturbation datasets.} Benchmark on a melanoma \citep{frangieh_multimodal_2021} and a leukemia \citep{papalexi_characterizing_2021} cell line. We report MapPFN (1) pre-trained on synthetic data and (2) fine-tuned on real data. MapPFN improves perturbation predictions across biological contexts by combining synthetic pre-training with fine-tuning on real single-cell data. The Mean and Linear baselines predict a single bulk profile per perturbation, so distributional metrics do not apply ($-$). Mean $\pm$ std over five data splits (see Appendix~\ref{sec:data_split}), each averaged over ten resampling seeds. Bold indicates results within one standard deviation of the best.}
\label{tab:singlecell_benchmark}
\resizebox{\textwidth}{!}{%
\begin{tabular}{llcccccc}
\toprule
Dataset & Method & $\text{W}_2$ $\downarrow$ & MMD ($\times 10^{-3}$) $\downarrow$ & RMSE $\downarrow$ & PDS $\downarrow$ & MR $\rightarrow 1$ & AUPRC $\uparrow$ \\
\midrule
\multirow[t]{7}{*}{Melanoma}
 & CPA \citep{lotfollahi_predicting_2023} & $\textBF{15.97} \pm 0.82$ & $152.13 \pm 46.92$ & $0.13 \pm 0.01$ & $0.49 \pm 0.01$ & $0.71 \pm 0.07$ & $0.03 \pm 0.01$ \\
 & CondOT \citep{bunne_supervised_2022} & $22.19 \pm 0.91$ & $\textBF{6.78} \pm 0.33$ & $\textBF{0.10} \pm 0.00$ & $0.06 \pm 0.02$ & $0.04 \pm 0.00$ & $0.41 \pm 0.05$ \\
 & MFM \citep{atanackovic_meta_2025} & $20.75 \pm 1.05$ & $\textBF{6.50} \pm 0.46$ & $\textBF{0.09} \pm 0.01$ & $0.11 \pm 0.05$ & $0.11 \pm 0.01$ & $0.46 \pm 0.07$ \\
 & CellFlow \citep{klein_cellflow_2025} & $22.28 \pm 0.59$ & $\textBF{6.80} \pm 0.49$ & $\textBF{0.10} \pm 0.01$ & $0.40 \pm 0.02$ & $0.02 \pm 0.00$ & $0.20 \pm 0.06$ \\
 & STATE \citep{adduri_predicting_2025} & $20.79 \pm 0.83$ & $7.39 \pm 1.19$ & $\textBF{0.09} \pm 0.01$ & $0.17 \pm 0.18$ & $\textBF{0.95} \pm 0.05$ & $0.42 \pm 0.23$ \\
 & MapPFN (pre-trained) & $22.25 \pm 1.33$ & $63.34 \pm 17.79$ & $0.32 \pm 0.04$ & $0.22 \pm 0.04$ & $\textBF{1.01} \pm 0.05$ & $0.34 \pm 0.07$ \\
 & MapPFN (fine-tuned) & $21.05 \pm 0.62$ & $\textBF{6.57} \pm 0.56$ & $\textBF{0.09} \pm 0.01$ & $\textBF{0.02} \pm 0.01$ & $\textBF{0.98} \pm 0.01$ & $\textBF{0.62} \pm 0.02$ \\
 \cmidrule(lr){2-8}
 & Identity & $23.14 \pm 0.64$ & $7.38 \pm 0.70$ & $0.11 \pm 0.01$ & $0.50 \pm 0.01$ & $0.00 \pm 0.00$ & $0.03 \pm 0.00$ \\
 & Mean & $-$ & $-$ & $0.30 \pm 0.02$ & $0.50 \pm 0.00$ & $-$ & $0.03 \pm 0.01$ \\
 & Linear & $-$ & $-$ & $0.30 \pm 0.02$ & $0.44 \pm 0.03$ & $-$ & $0.09 \pm 0.06$ \\
 & Observed & $7.66 \pm 0.40$ & $1.96 \pm 0.10$ & $0.04 \pm 0.00$ & $0.00 \pm 0.00$ & $1.00 \pm 0.00$ & $0.82 \pm 0.02$ \\
\midrule
\multirow[t]{7}{*}{Leukemia}
 & CPA \citep{lotfollahi_predicting_2023} & $\textBF{16.64} \pm 3.40$ & $112.31 \pm 29.79$ & $0.23 \pm 0.11$ & $0.51 \pm 0.01$ & $0.76 \pm 0.28$ & $\textBF{0.11} \pm 0.03$ \\
 & CondOT \citep{bunne_supervised_2022} & $22.93 \pm 2.12$ & $29.21 \pm 10.51$ & $0.31 \pm 0.07$ & $0.51 \pm 0.02$ & $0.28 \pm 0.24$ & $\textBF{0.10} \pm 0.05$ \\
 & MFM \citep{atanackovic_meta_2025} & $22.12 \pm 11.41$ & $23.76 \pm 8.54$ & $0.24 \pm 0.13$ & $0.52 \pm 0.03$ & $0.82 \pm 0.32$ & $\textBF{0.11} \pm 0.04$ \\
 & CellFlow \citep{klein_cellflow_2025} & $\textBF{19.76} \pm 4.52$ & $\textBF{12.99} \pm 5.51$ & $\textBF{0.18} \pm 0.05$ & $0.52 \pm 0.02$ & $0.04 \pm 0.03$ & $\textBF{0.11} \pm 0.04$ \\
 & STATE \citep{adduri_predicting_2025} & $\textBF{18.47} \pm 4.34$ & $\textBF{12.25} \pm 5.60$ & $\textBF{0.15} \pm 0.05$ & $0.50 \pm 0.01$ & $0.87 \pm 0.06$ & $\textBF{0.12} \pm 0.04$ \\
 & MapPFN (pre-trained) & $36.23 \pm 8.24$ & $132.95 \pm 60.60$ & $0.61 \pm 0.18$ & $\textBF{0.47} \pm 0.02$ & $1.95 \pm 0.82$ & $\textBF{0.13} \pm 0.05$ \\
 & MapPFN (fine-tuned) & $\textBF{19.90} \pm 4.37$ & $\textBF{12.89} \pm 4.37$ & $\textBF{0.17} \pm 0.04$ & $\textBF{0.41} \pm 0.06$ & $\textBF{0.96} \pm 0.08$ & $\textBF{0.14} \pm 0.05$ \\
 \cmidrule(lr){2-8}
 & Identity & $19.41 \pm 4.36$ & $11.63 \pm 5.70$ & $0.14 \pm 0.06$ & $0.51 \pm 0.00$ & $0.00 \pm 0.00$ & $0.12 \pm 0.04$ \\
 & Mean & $-$ & $-$ & $0.17 \pm 0.04$ & $0.50 \pm 0.00$ & $-$ & $0.11 \pm 0.04$ \\
 & Linear & $-$ & $-$ & $0.20 \pm 0.03$ & $0.50 \pm 0.04$ & $-$ & $0.11 \pm 0.04$ \\
 & Observed & $12.68 \pm 2.42$ & $2.96 \pm 0.40$ & $0.05 \pm 0.00$ & $0.04 \pm 0.01$ & $1.00 \pm 0.00$ & $0.86 \pm 0.07$ \\
\bottomrule
\vspace{-1.5em}
\end{tabular}}
\end{table*}

\paragraph{Interventional context improves performance over observational data alone} We evaluate whether MapPFN benefits from improved identifiability by conditioning on interventional data. Specifically, we ablate the effect of providing a set of interventional distributions $\mathcal{C}$ versus the observational-only setting, where $\mathcal{C} = \emptyset$. As shown in \autoref{tab:frangieh_ablation}, conditioning on interventional distributions improves performance across all metrics over using only observational data. Since the model architecture remains unchanged, this suggests that interventional context enables MapPFN to learn perturbation-specific mappings not accessible from observational data alone. Performance further improves with the number of interventional experiments provided in context (see \autoref{fig:context_sizes} in \autoref{sec:additional_results}).

\paragraph{Counterfactual paired prior improves downstream performance} To isolate the task of causal inference from the additional difficulty introduced by unpaired data, we follow \citet{robertson_do-pfn_2025} and pre-train MapPFN on counterfactual interventional data, achieved by keeping the random seed of SERGIO constant across treatments. This ensures that the differences between interventional distributions are not driven by a difference in initial condition to the stochastic differential equation, but only by the differences in underlying mechanism and perturbation effects. \autoref{fig:prior_convergence} shows the Pearson correlation between the feature-wise variances of the predicted and ground-truth post-perturbation distribution on the validation set, evaluated separately for the paired and unpaired prior. The paired prior converges to a variance correlation of approximately 0.8 within 50k steps, while the unpaired prior saturates around 0.6 even after 400k steps. The paired prior results in a substantial improvement across all metrics except PDS on real single-cell data (\autoref{tab:frangieh_ablation}). We hypothesize that counterfactual interventional distributions provide stronger signal by isolating causal effects from the added variability of unpaired samples.

\begin{table*}[htbp]
\centering
\caption{\textbf{Ablation of the counterfactual paired prior, interventional context and synthetic pre-training on the melanoma dataset.} Removing the counterfactual prior replaces paired data with unpaired interventional distributions. Removing the interventional context implies conditioning only on observational data. Random init trains the same architecture from scratch on real data, without synthetic pre-training. Mean $\pm$ std over five data splits, each averaged over ten resampling seeds.}
\label{tab:frangieh_ablation}
\resizebox{0.9\textwidth}{!}{%
\begin{tabular}{lcccccc}
\toprule
Configuration & $\text{W}_2$ $\downarrow$ & MMD ($\times 10^{-3}$) $\downarrow$ & RMSE $\downarrow$ & PDS $\downarrow$ & MR $\rightarrow 1$ & AUPRC $\uparrow$ \\
\midrule
MapPFN (pre-trained) & $22.25 \pm 1.33$ & $63.34 \pm 17.79$ & $0.32 \pm 0.04$ & $0.22 \pm 0.04$ & $1.01 \pm 0.05$ & $0.34 \pm 0.07$ \\
\quad $-$ counterfactual prior & $69.15 \pm 7.82$ & $143.51 \pm 7.78$ & $0.94 \pm 0.06$ & $0.18 \pm 0.02$ & $3.16 \pm 0.37$ & $0.08 \pm 0.03$ \\
\quad $-$ interventional context & $40.99 \pm 4.93$ & $101.05 \pm 9.67$ & $0.65 \pm 0.06$ & $0.29 \pm 0.04$ & $1.85 \pm 0.18$ & $0.25 \pm 0.05$ \\
\midrule
MapPFN (fine-tuned) & $21.05 \pm 0.62$ & $6.57 \pm 0.56$ & $0.09 \pm 0.01$ & $0.02 \pm 0.01$ & $0.98 \pm 0.01$ & $0.62 \pm 0.02$ \\
MapPFN (random init) & $24.34 \pm 0.90$ & $52.33 \pm 3.49$ & $0.31 \pm 0.02$ & $0.19 \pm 0.06$ & $1.13 \pm 0.02$ & $0.09 \pm 0.03$ \\
\bottomrule
\end{tabular}}
\end{table*}

\paragraph{MapPFN scales to larger gene sets and number of cells at inference time} Pre-trained on \emph{in silico} knockouts, MapPFN uniquely adapts to new gene sets without retraining. To scale beyond the 50 genes seen during training, we apply test-time augmentation (TTA). We sample random overlapping subsets of 50 genes from 100, predict cell-level post-perturbation distributions for each subset, pool predicted cells per gene across subsets, and identify differentially expressed genes via per-gene statistical testing. As shown in \autoref{fig:tta}, TTA improves AUPRC, indicating that MapPFN scales to larger gene sets than seen during pre-training. Performance also improves beyond the training configuration with more cells per perturbation in context, showing that MapPFN adapts predictions to the data via in-context learning (see \autoref{fig:cell_scaling} in~\autoref{sec:additional_results}).

\section{Discussion}

We introduced MapPFN, the first prior-data fitted network for perturbation prediction, framing the problem as a distribution mapping with a multi-experiment interventional context. During pre-training, MapPFN meta-learns to map between pre- and post-perturbation distributions from a synthetic biological prior. At inference time, it predicts perturbation effects in new biological contexts through in-context learning.

In a controlled linear SCM benchmark where the causal mechanism is known, MapPFN successfully meta-learns perturbation prediction, demonstrating the feasibility of our framing. Evaluating MapPFN on single-cell perturbation datasets, we find that combining synthetic pre-training with fine-tuning on real data improves perturbation predictions across biological contexts. Our ablations show that conditioning on interventional experiments and using a counterfactual paired prior each improve performance.

MapPFN tackles two key challenges in perturbation prediction: (1) decoupling pre-training from the limited availability of experimental perturbation datasets through the synthetic biological prior, and (2) adapting a single pre-trained model to new biological contexts and gene sets without retraining. These capabilities distinguish MapPFN from existing perturbation models that are constrained by the size of available perturbation datasets and require retraining for each new dataset or gene set.

\paragraph{Limitations} MapPFN depends on the synthetic prior to generalize across biological contexts, requiring further systematic evaluation of prior design choices. The synthetic biological prior models hard genetic knockouts, restricting evaluation to CRISPR knockout screens. Extending it to support soft knockdowns induced by CRISPRi-based screens, such as those in the Virtual Cell Challenge \citep{roohani_virtual_2025}, is a natural next step. Extending MapPFN to support combinatorial, drug-based or chemical perturbations remains an open challenge \citep{schneider_generative_2025, dong_stack_2026, wu_identifying_2025}. Finally, enforcing permutation equivariance over genes, scaling MapPFN to higher-dimensional input spaces \citep{kolberg_tabpfn-wide_2025} and improving simulation efficiency \citep{chatzaroulas_sergio_rs_2024} to model larger systems offer promising directions.

\paragraph{Conclusion} Given the success of PFNs in tabular prediction and causal inference, we are optimistic that scaling MapPFN in terms of model capacity and extending the synthetic prior to more diverse regulatory mechanisms and broader perturbation types will yield further improvements. Our findings suggest that meta-learning perturbation prediction on synthetic biological priors with test-time interventional context offers a scalable path toward context-adaptive virtual cell foundation models.

\begin{ack}
We thank Michael Plainer, Jonas Loos, Alexander Möllers and Lukas Ruff for the fruitful discussions and helpful input. This work was supported by the German Federal Ministry of Research, Technology and Space under the grant BIFOLD26B. Marvin Sextro and Gabriel Dernbach were employed part-time at Aignostics during this project.
\end{ack}

\bibliographystyle{unsrtnat}
\bibliography{neurips_2026}


\newpage
\appendix

\section*{Appendix}
\vspace{-1em}
\startcontents[appendix]
\printcontents[appendix]{}{1}{\setcounter{tocdepth}{2}\medskip}

\section{Model} \label{sec:model}

\subsection{Architecture}

We build on the multimodal diffusion transformer (MMDiT) \citep{esser_scaling_2024} architecture from the Stable Diffusion 3 family. Instead of text and image modalities, we keep the denoising process, the pre- and post-treatment data as well as the treatment in three modality streams. With this setup, each modality has separate weights and information flows between modalities via joint attention. As we are working with sets of cells, we use the permutation invariance of the attention mechanism by removing the sinusoid positional encoding. Instead, we add learnable embeddings (a) for each treatment in the context to tell apart different conditions, (b) to tell apart observational and interventional data, and (c) to tell apart the query condition from the context. Our model has 8 layers with an embedding dimension of 256 and a $2\times$ expansion to 512 in the feed-forward layers. We append 8 register tokens to the noise stream and use 4 multi-head attention heads of size 64 each. Time conditioning is implemented by feature-wise linear modulation (FiLM) \citep{perez_film_2018}. Overall, this configuration amounts to approximately 26M trainable parameters.

MMDiT is an off-the-shelf transformer architecture established for visual generation that supports generating high-dimensional distributions and in-context learning via joint attention. We choose it over the Diffusion Transformer (DiT) \citep{peebles_scalable_2023} for its reported improvement on multimodal data \citep{esser_scaling_2024}, as the context of MapPFN contains mixed modalities in the form of one-hot perturbation identifiers, data and noise inputs. MMDiT differs from DiT in maintaining separate weights per modality while attending jointly across them, rather than sharing a single set of weights across all tokens. We ablate this choice in Appendix~\ref{sec:architecture_ablation}.

\subsection{Pre-training}

We pre-train our model using a flow matching \citep{lipman_flow_2023} objective with an affine Gaussian probability path. During training, we randomly drop the condition by replacing it with a learnable \texttt{null} embedding with probability $p=0.2$. Following \citet{esser_scaling_2024}, we sample $t\sim \text{LogitNormal}(0,1)$. We use the AdamW optimizer \citep{loshchilov_decoupled_2019} with a warmup-stable-decay learning rate schedule \citep{dremov_training_2025} using 1\% of the total number of steps for warmup to a peak learning rate of $10^{-4}$ and 20\% for a square root decay. We maintain an exponential moving average (EMA) of model weights with a decay of $0.999$ and use these weights for inference \citep{esser_scaling_2024}. We pre-train for 50k steps on the linear SCM prior and 400k steps on the synthetic biological prior.

\subsection{Fine-tuning}

We fine-tune MapPFN for 3,000 iterations with a linear warmup to a reduced peak learning rate of $5 \times 10^{-5}$, taking approximately 10 minutes on a single GPU.

\subsection{Inference}

To generate samples, we integrate the learned flow by solving its ordinary differential equation (ODE) using the Dopri5 \citep{dormand_family_1980} solver, as implemented in \texttt{diffrax} \citep{kidger_neural_2022}. We use classifier-free guidance \citep{ho_classifier-free_2021} for conditional generation with a guidance weight $\omega = 2.0$ by default.

\section{Priors}

\subsection{Structural Causal Models} \label{sec:synthetic_scm_data}

\paragraph{Linear Additive Noise Models} We generate synthetic observational and interventional data using a linear additive noise model (ANM) with Gaussian noise of the form $\mathbf{z} = \mathbf{Wz} + \bm{\epsilon}$, where $\mathbf{W} \in \mathbb{R}^{d \times d}$ is a weighted adjacency matrix encoding the causal graph and $\bm{\epsilon} \sim \mathcal{N}(0, \mathbf{I})$ represents independent additive noise. The underlying directed acyclic graph (DAG) is sampled from an Erd\H{o}s-R\'{e}nyi \citep{erdos_evolution_1960} model $\mathcal{G}(d, p)$ with $d=20$ nodes and an edge probability of $p = 0.5$, restricted to the upper triangular structure under a random node permutation to ensure acyclicity. Edge weights are sampled uniformly from $[-2, -0.5] \cup [0.5, 2]$, ensuring coefficients are bounded away from zero to exclude negligible causal effects. To ensure observations have approximately unit variance and fall within the $[-2,2]$ range, we normalize the weight matrix by rescaling $\mathbf{W} \leftarrow \mathbf{D}^{-1/2} \mathbf{W}$ where $\mathbf{D} = \operatorname{diag}(\mathbf{TT}^\top)$ and $\mathbf{T} = (\mathbf{I} - \mathbf{W})^{-1}$ denotes the transfer matrix. To avoid varsortability \citep{reisach_beware_2021}, we scale all variables of the generated data to unit variance.

\paragraph{Atomic Interventions} Interventional data is generated following Pearl's do-calculus: for an intervention $\operatorname{do}(t)$, we remove all incoming edges to the intervened node and set its value to $c \sim \text{Unif}([0.5,1.5])$, simulating a gene perturbation experiment where the treated genes have varying perturbation efficiencies. To condition the model on the treatment, we use a $d$-dimensional one-hot-encoding, where the element at the hot index contains the intervention value $c$.

\paragraph{Experiment Design} We intervene on each of the 20 nodes in 1000 randomly generated DAGs to generate all 20k possible context/treatment conditions. Per treatment condition, we sample $n=500$ pre-perturbation observations, resulting in 10M interventional vector-valued samples. Additionally, we generate 500 untreated observations per DAG, adding to a total of 10.5M samples. For MapPFN, we use a context $\mathcal{C}$ with $K=4$ perturbation experiments.

\subsection{Synthetic Biological Prior} \label{sec:synthetic_rna_data}

To generate synthetic perturbation datasets across diverse contexts, we combine a preferential attachment algorithm for sampling graphs with properties close to real GRNs \citep{aguirre_gene_2025} and SERGIO \citep{dibaeinia_sergio_2020} for simulating observations from these graphs using Hill functions and adding technical noise. Our goal is maximally broad but relevant prior coverage. We sample from a family of SERGIO settings validated across 15 real datasets \citep{dibaeinia_sergio_2020} and exclude technical noise configurations that do not match the 10x Chromium sequencing protocol, as we found these had the largest impact on distributional similarity.

\paragraph{Gene Regulatory Networks} GRNs have unique properties that we want the prior to replicate. As summarized by \citet{aguirre_gene_2025}, these properties are (1) sparsity, (2) directed edges and cycles, (3) asymmetry of in- and out-degree distributions and (4) modularity. To ensure our dataset captures the diversity of GRNs, we sample the hyperparameters uniformly from ranges suggested by \citet{aguirre_gene_2025}, as summarized in \autoref{tab:grn_params}.

Since SERGIO requires GRNs that are acyclic, we remove cycles by removing the edge with the smallest absolute weight in each cycle. Additionally, SERGIO requires at least one master regulator, i.e. genes with no incoming edges but at least one outgoing edge. If no master regulators exist after cycle removal, we select the top 5\% of genes with the lowest in-degree among all genes with outgoing edges and remove all incoming edges, forcing them to become master regulators.

\paragraph{Simulation} Given a regulatory network sampled in the previous step, we simulate single-cell expression profiles using SERGIO \citep{dibaeinia_sergio_2020}. SERGIO models the expression level of each gene as a function of its regulators using Hill functions \citep{gesztelyi_hill_2012}. It then models the gene interaction dynamics by solving a Stochastic Differential Equation (SDE) called chemical Langevin equation (CLE) \citep{gillespie_chemical_2000}. Single-cell expression values are generated by applying technical noise to the steady state of this system. We sample the hyperparameters for the simulation and technical noise uniformly from the ranges summarized in \autoref{tab:sergio_params}. For improved simulation speed, we use a reimplementation of SERGIO in Rust \citep{chatzaroulas_sergio_rs_2024}.

\paragraph{Experiment Design} We sample single-cell data in 6000 synthetic GRNs of 50 genes and simulate $n=200$ single-cell expression profiles per treatment condition. We use a context $\mathcal{C}$ containing $K=8$ perturbation experiments.

\begin{table}
\centering
\caption{GRN structure parameters for the graph generator.}
\label{tab:grn_params}
\begin{tabular}{lll}
\toprule
\textbf{Symbol} & \textbf{Description} & \textbf{Range} \\
\midrule
$k$ & Number of gene groups/modules & $\{1, 2, 3\}$ \\
$p$ & Sparsity term (avg. regulators per gene) & $[1.5, 3.0]$ \\
$\delta_{\text{in}}$ & In-degree uniformity term & $[10, 300]$ \\
$\delta_{\text{out}}$ & Out-degree uniformity term & $[1, 30]$ \\
$w$ & Modularity term (within-group connectivity) & $[1, 900]$ \\
\bottomrule
\end{tabular}
\end{table}

\begin{table}
\centering
\caption{SERGIO simulation and technical noise parameters.}
\label{tab:sergio_params}
\begin{tabular}{lll}
\toprule
\textbf{Symbol} & \textbf{Description} & \textbf{Range} \\
\midrule
\multicolumn{3}{l}{\textit{Simulation parameters}} \\
$k$ & Interaction strengths & $[1.0, 5.0]$ \\
$b$ & Master regulator production rates & $[0.5, 2.0] \cup [3.0, 5.0]$ \\
$\gamma$ & Hill function coefficients (nonlinearity) & $[1.5, 2.5]$ \\
$\lambda$ & Decay rates per gene & $[0.5, 1.0]$ \\
$\zeta$ & Stochastic process noise scale & $[0.5, 1.5]$ \\
\midrule
\multicolumn{3}{l}{\textit{Technical noise parameters}} \\
$\mu_{\text{outlier}}$ & Log-normal outlier mean & $[0.8, 5.0]$ \\
$\mu_{\text{lib}}$ & Log-normal library size mean & $[4.5, 6.0]$ \\
$\sigma_{\text{lib}}$ & Log-normal library size std & $[0.3, 0.7]$ \\
$\delta$ & Dropout percentile & $[8.0, 8.0]$ \\
$\xi$ & Dropout temperature & $[45.0, 82.0]$ \\
\bottomrule
\end{tabular}
\end{table}

\section{Single-cell Perturbation Datasets} \label{sec:single_cell_data}

We obtain two single-cell perturbation datasets from \texttt{pertpy} \citep{heumos_pertpy_2025}. Both use CRISPR knockout perturbations, matching the hard interventions modeled by SERGIO. To keep the synthetic prior grounded in a well-understood simulator validated against real gene expression data \citep{dibaeinia_sergio_2020}, we restrict evaluation to datasets compatible with this intervention type. CRISPRi-based datasets, such as those in the Virtual Cell Challenge \citep{roohani_virtual_2025}, induce soft knockdowns and require extending the simulator to support partial gene suppression.

\subsection{Melanoma Dataset} The first dataset \citep{frangieh_multimodal_2021} contains approximately 218,000 cells measured using Perturb-CITE-seq under 248 CRISPR gene knockout perturbations. Perturbed genes were selected by their membership in an immune evasion program associated with resistance to immunotherapy. The knockout perturbations were measured in a patient-derived melanoma cell line under three treatment-induced cell states: one treated with interferon-$\gamma$ (IFN-$\gamma$) to put the cells into an alarmed state, a co-culture treated with tumor infiltrating lymphocytes (TIL) to simulate an immune response, and an untreated control. For our experiments, we use the IFN-$\gamma$ state as the holdout context.

\subsection{Leukemia Dataset} The second dataset \citep{papalexi_characterizing_2021} was generated using ECCITE-seq, a multimodal assay combining scRNA-seq with surface protein measurements, on the THP-1 monocytic leukemia cell line. Cells were stimulated with IFN-$\gamma$, decitabine (DAC) and TGF-$\beta$1 to induce PD-L1 expression. CRISPR perturbations target 25 genes involved in immune checkpoint regulation. After quality control and assignment to a single perturbation, approximately 20,000 cells are available for analysis. Unlike the melanoma dataset, this dataset contains a single cell state.

\subsection{Preprocessing} \label{sec:preprocessing} Following best practice for single-cell RNA sequencing preprocessing \citep{luecken_current_2019}, we first normalize the total counts per cell to be equal to the median total count across all cells, followed by a \texttt{log1p} transform
\begin{align}
   \mathbf{\tilde x} = \log_2\left(1 + \frac{m \cdot \mathbf{x}}{\|\mathbf{x}\|_1}\right)
\end{align}
where $m = \operatorname{median}_i(\|\mathbf{x}_i\|_1)$ is the median total count across all cells. We use the implementation of \texttt{sc.pp.normalize\_total} and \texttt{sc.pp.log1p} provided by \texttt{scanpy} \citep{wolf_scanpy_2018}. We retain perturbations with at least 200 cells. For both datasets, we select the set of perturbed genes and complete the set to 50 genes with the top marker genes, identified by differential expression analysis between each perturbation and control on the training conditions of each split using \texttt{sc.tl.rank\_genes\_groups}. For the melanoma dataset, all 50 genes can be selected from the 248 perturbation targets, which belong to a shared immune evasion program \citep{frangieh_multimodal_2021}. For the leukemia dataset, only 23 of the 25 perturbation targets remain after filtering, and the other 27 genes are marker genes that are not themselves perturbation targets. We sample $n=200$ i.i.d.\ cells per condition.

\section{Experimental Details} \label{sec:experimental_details}

\subsection{Data Split} \label{sec:data_split}

\begin{figure}
  \vskip 0.2in
  \begin{center}
    \centerline{\includegraphics[width=0.2\columnwidth]{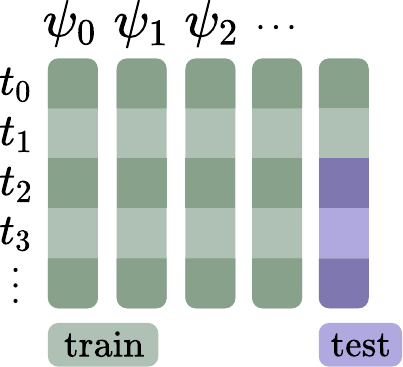}}
    \caption{
      Data split. Each box represents a dataset $\Yint_{ij} \in \mathbb{R}^{N\times d}$ sampled from the SCM $\psi_i$ under treatment $t_j$. Green boxes are part of the training data and purple boxes are withheld for evaluation.
    }
    \label{fig:data_split}
  \end{center}
  \vskip -0.2in
\end{figure}

We split the data at the condition level, where each condition corresponds to a context-treatment pair $(\psi_i, t_j)$. Each pair is assigned independently to the train, validation, or test split, ensuring that the samples of a particular context/treatment condition are only contained in a single split. By default, we report the mean and standard deviation over five condition-level splits, while \autoref{fig:inference_scaling} and \autoref{fig:prior_coverage} use a single split. Earlier versions of this work selected marker genes using all conditions and reported results on a single condition-level split. Since both the held-out perturbations and the selected gene set depend on the split, performance varies across splits. As shown in \autoref{fig:data_split}, a random half of the treatments of the holdout context are assigned to the test split, while the remainder is included in the train split. Following the Virtual Cell Challenge~\cite{roohani_virtual_2025}, the training data includes interventional distributions from a subset of perturbations in the holdout context. MapPFN makes predictions only using this subset as interventional context, while all baselines are trained in a supervised manner on the full training set, including the subset of perturbations in the holdout context. To obtain a validation set, we randomly select 10\% of the remaining training conditions.

\subsection{Baselines} \label{sec:baselines}

For comparability, MapPFN and all baselines are conditioned on one-hot encoded treatments.

\paragraph{CPA} The Compositional Perturbation Autoencoder (CPA) \citep{lotfollahi_predicting_2023} decomposes cell states into independent basal state, treatment, and covariate embeddings using a variational autoencoder. Perturbation effects are modeled as additive shifts in latent space, enabling combinatorial generalization to unseen treatment combinations.

\paragraph{CondOT} Conditional Optimal Transport (CondOT) trains a partially input-convex neural network (PICNN) \citep{amos_input_2017} to learn a global conditional OT map for different treatment conditions or subpopulations \citep{bunne_supervised_2022}. We use the identity initialization, as the Gaussian initialization requires target distribution statistics that are unavailable for unseen contexts.

\paragraph{Meta Flow Matching} Meta Flow Matching (MFM) \citep{atanackovic_meta_2025} proposes to integrate the vector fields on the Wasserstein manifold by conditioning the flow on a learned representation of the observational distribution. With the aim of modeling interactions between individual cells, MFM separately trains a graph neural network (GNN) yielding population embeddings.

\paragraph{CellFlow} CellFlow \citep{klein_cellflow_2025} uses flow matching to learn a transport from the source to the perturbed cell distribution. Perturbation covariates are encoded into a condition embedding that guides the flow. To handle arbitrary numbers of perturbations in a permutation-invariant manner, CellFlow employs set aggregation with multihead attention.

\paragraph{STATE} STATE \citep{adduri_predicting_2025} consists of a State Transition model (ST) that predicts perturbation effects on sets of cells, and an optional State Embedding model (SE) that provides pre-trained cell representations from large-scale observational data. ST uses self-attention over sets of observational cells to predict perturbed cell populations, modeling interactions between cells within each set. Since our gene sets are low-dimensional, we use the ST model on raw expression profiles.

\paragraph{Naive baselines} Following \citet{ahlmann-eltze_deep-learning-based_2025}, Mean predicts the average of the per-perturbation pseudobulk profiles in the training split. Linear is a rank-constrained linear model over PCA-derived gene and perturbation embeddings (rank 8, ridge penalty $\lambda = 0.1$). Both are fit on the same data split and gene set as all other methods.

\subsection{Metrics} \label{sec:metrics}

We measure the discrepancy between the distribution of predicted samples and the distribution of ground-truth samples. We evaluate our models in terms of distributional, correlation and ranking-based metrics.

\paragraph{Wasserstein Distance} The entropy-regularized Wasserstein distance \citep{cuturi_sinkhorn_2013} between ground-truth samples $\mathbf{Y} \in \mathbb{R}^{n \times d}$ and predicted samples $\hat{\mathbf{Y}} \in \mathbb{R}^{m \times d}$ is computed as
\begin{equation}
    W_2(\mathbf{Y}, \hat{\mathbf{Y}}) := \left(\min_{\mathbf{P} \in \mathcal{U}(\mathbf{Y}, \hat{\mathbf{Y}})} \sum_{i=1}^n \sum_{j=1}^m \mathbf{P}_{ij} ||\mathbf{y}_i - \hat{\mathbf{y}}_j||^2_2 - \epsilon H(\mathbf{P})\right)^{1/2}
\end{equation}
where $\epsilon$ is the regularization parameter, $\mathcal{U}(\mathbf{Y}, \hat{\mathbf{Y}})$ is the set of transport matrices of shape $n \times m$ given by
\begin{equation}
    \mathcal{U} = \left\{\mathbf{P} \in \mathbb{R}^{n \times m}_{\geq 0} \colon \mathbf{P}\mathbf{1}_m = \frac{1}{n} \cdot \mathbf{1}_n \text{ and } \mathbf{P}^\top \mathbf{1}_n = \frac{1}{m} \cdot \mathbf{1}_m\right\}
\end{equation}
and $H$ is the entropy computed as $H(\mathbf{P}) = -\sum_{ij} \mathbf{P}_{ij} \log \mathbf{P}_{ij} - 1$. To obtain a valid distance that becomes zero if and only if the compared distributions are equal, we use the Sinkhorn divergence \citep{feydy_interpolating_2019, genevay_learning_2018} given by
\begin{equation}
    S_2(\mathbf{Y}, \hat{\mathbf{Y}}) = W_2(\mathbf{Y}, \hat{\mathbf{Y}}) - \frac{1}{2} W_2(\mathbf{Y}, \mathbf{Y}) - \frac{1}{2} W_2(\hat{\mathbf{Y}}, \hat{\mathbf{Y}})
\end{equation}
We use the implementation provided by the optimal transport tools (OTT) package  \citep{cuturi_optimal_2022} with the regularization parameter $\epsilon = 0.1$.

\paragraph{Maximum Mean Discrepancy}

The squared maximum mean discrepancy (MMD) \citep{gretton_kernel_2012} between ground truth and predicted samples $\mathbf{Y}$ and $\mathbf{\hat{Y}}$ for a conditionally positive definite kernel $k$ is defined as
\begin{equation}
    \operatorname{MMD}^2(\mathbf{Y}, \mathbf{\hat{Y}}) = \mathbb{E}_{\mathbf{y},\mathbf{y}'}[k(\mathbf{y},\mathbf{y}')] + \mathbb{E}_{\mathbf{\hat{y}}, \mathbf{\hat{y}}'}[k(\mathbf{\hat{y}}, \mathbf{\hat{y}}')] - 2 \mathbb{E}_{\mathbf{y},\mathbf{\hat{y}}}[k(\mathbf{y},\mathbf{\hat{y}})]
\end{equation}

We compute the MMD for the Gaussian radial basis function (RBF) kernel
\begin{equation}
    k_\text{RBF}(\mathbf{x}, \mathbf{y}) = \exp{\left(-\gamma ||\mathbf{x} - \mathbf{y}||_2^2\right)}
\end{equation}
and report the mean over multiple length scales $\gamma \in \{10, 1, 0.1, 0.01, 0.001\}$.

\paragraph{Root Mean Squared Error} We follow \citet{wu_perturbench_2025} in computing the root mean squared error (RMSE) 
\begin{equation}
    \operatorname{RMSE}(\mathbf{Y}, \hat{\mathbf{Y}}) = \sqrt{\frac{1}{n}\sum_i^n \left(\hat{\mu}_i - \mu_i\right)^2}
\end{equation}
between the mean of the predicted and ground-truth post-perturbation distributions $\bm{\mu} = \mathbb{E}[\mathbf{y}^\text{int}]$ and $\bm{\hat{\mu}} = \mathbb{E}[\mathbf{\hat{y}}^\text{int}]$.

\paragraph{Perturbation Discrimination Score} To evaluate whether model predictions are distinguishable across perturbations, we adopt the perturbation discrimination score (PDS) from \citet{wu_perturbench_2025}. Let $\bm{\mu}_i = \mathbb{E}[\mathbf{y}^\text{int}_i]$ and $\bm{\hat{\mu}}_i = \mathbb{E}[\hat{\mathbf{y}}^\text{int}_i]$ denote the mean observed and predicted expression for perturbation $i$, respectively. The PDS measures, for each perturbation $i$, what fraction of other observations $\bm{\mu}_j$ are closer to $\bm{\hat{\mu}}_i$ than the matched observation $\bm{\mu}_i$:
\begin{equation}
    \operatorname{PDS}_{\text{avg}} = \frac{1}{p} \sum_{i=1}^{p} \operatorname{PDS}(\bm{\hat{\mu}}_i), \quad \operatorname{PDS}(\bm{\hat{\mu}}_i) = \frac{1}{p-1} \sum_{\substack{1 \leq j \leq p \\ j \neq i}} \mathbb{I}\left( d(\bm{\hat{\mu}}_i, \bm{\mu}_j) \leq d(\bm{\hat{\mu}}_i, \bm{\mu}_i) \right)
\end{equation}
where $p$ is the number of perturbations and $d$ is the Euclidean distance. This metric ranges from 0 (perfect) to 1 (worst), with 0.5 corresponding to random predictions. The PDS is particularly sensitive to mode collapse, as a model generating similar predictions for all perturbations will have many ground-truth observations closer than the matched one.

\paragraph{Area Under the Precision Recall Curve} To evaluate whether model predictions reliably imply identification of differentially expressed genes (DEGs), we adopt the AUPRC metric from \citet{zhu_auprc_2025}. For a given perturbation, ground-truth DEGs are identified using a per-gene Wilcoxon rank-sum test comparing single-cell expression values before and after intervention, under the null hypothesis of identical distributions \citep{wilcoxon_individual_1945}. Benjamini-Hochberg \citep{benjamini_adaptive_2000} correction is applied across genes, and DEGs are defined by jointly thresholding on effect size and statistical certainty, using the absolute $\log_2$ fold-change ($\tau_l = 0.2$) and the negative $\log_{10}$ p-value ($\tau_p = 2$).
\begin{equation}
    Z_g = \mathbb{I}\left(\tilde{p}_g > \tau_p \land |\tilde{l}_g| > \tau_l\right)
\end{equation}
where $\tilde{p}_g = -\log_{10}(p_g)$ and $\tilde{l}_g = \log_2(\tilde{\mu}_g^\text{int} / \tilde{\mu}_g^\text{obs})$ denote the negative log p-value and log fold-change for gene $g$, respectively. For in silico predictions, we compute a ranking score $R_g = |\hat{l}_g| \cdot \mathbb{I}(\hat{p}_g > \tau_p)$ that combines the magnitude of predicted expression change with statistical significance. By varying a threshold $r$ on this score, we generate a family of classifiers $\hat{Z}_g(r) = \mathbb{I}(R_g > r)$ and construct precision-recall curves against the ground-truth labels $Z_g$, summarized by the AUPRC. Differential expression analysis was performed using \texttt{scanpy.tl.rank\_genes\_groups} \citep{wolf_scanpy_2018}.

\begin{table}[t]
\caption{Hyperparameter search ranges for each method.}
\label{tab:hyperparameters}
\begin{center}
\begin{tabular}{llr}
\toprule
\textbf{Method} & \textbf{Hyperparameter} & \textbf{Search Range} \\
\midrule
MapPFN & Classifier-free guidance weight & $\{1.0, 1.5, 2.0, 2.5, 3.0\}$ \\
\midrule
CPA    & \multicolumn{2}{l}{Following the \href{https://github.com/altoslabs/perturbench/blob/4825e392294768da4b35561a76502c7006d6453e/src/perturbench/configs/hpo/cpa_hpo.yaml}{tuning protocol} of PerturBench~\citep{wu_perturbench_2025}.} \\         
\midrule
CondOT & Hidden dimensions & $\{64, 128, 256\}$ \\
       & Hidden layers & $\{2, 3, 4\}$ \\
       & Learning rate & $\{10^{-3}, 10^{-4}, 10^{-5}\}$ \\
\midrule
MFM & k-nearest neighbors & $\{0, 10, 50, 100\}$ \\
       & GNN embedding dimensions & $\{64, 128, 256\}$ \\
       & Learning rate & $\{10^{-3}, 10^{-4}, 10^{-5}\}$ \\
\midrule
CellFlow & \multicolumn{2}{l}{Following the \href{https://github.com/theislab/CellFlow/blob/03bff12e71326742ad53a0909a4a8fee36958794/docs/notebooks/500_combosciplex.ipynb}{reference notebook}~\citep{klein_cellflow_2025}.} \\
\midrule
STATE  & \multicolumn{2}{l}{Following the \href{https://colab.research.google.com/drive/1Ih-KtTEsPqDQnjTh6etVv_f-gRAA86ZN}{reference notebook}~\citep{adduri_predicting_2025}.} \\
\bottomrule
\end{tabular}
\end{center}
\end{table}

\subsection{Hyperparameters}

By default, we use the hyperparameters recommended by the authors of each baseline. We follow the Optuna-based tuning protocol of PerturBench~\citep{wu_perturbench_2025} for CPA, perform a small grid search for CondOT and MFM, and use the published reference configurations for STATE~\citep{adduri_predicting_2025} and CellFlow~\citep{klein_cellflow_2025}. For MapPFN, we only grid-search the classifier-free guidance weight. The searched hyperparameters are summarized in \autoref{tab:hyperparameters}.

\subsection{Implementation} \label{sec:implementation}

We use \texttt{JAX} \citep{bradbury_jax_2018} to implement our experiments. Our model is implemented using \texttt{equinox} \citep{kidger_equinox_2021} and \texttt{diffrax} \citep{kidger_neural_2022} for ODE solving. We also make use of Optimal Transport Tools (\texttt{OTT}) \citep{cuturi_optimal_2022} to compute the Sinkhorn distance. We use \texttt{hydra-zen} \citep{soklaski_tools_2022} to configure our experiments. For single-cell data processing, we build upon the \texttt{scverse} ecosystem, including \texttt{anndata} \citep{virshup_anndata_2024}, \texttt{scanpy} \citep{wolf_scanpy_2018} and \texttt{pertpy} \citep{heumos_pertpy_2025}.

We run our experiments on a high-performance cluster, using a single NVIDIA A100 or H100 GPU with 80 GB of VRAM for training. For the linear SCM dataset, each experiment ran for 2-8h depending on the method and configuration. Pre-training MapPFN on synthetic single-cell data took approximately 10-36h, depending on the setting and corresponding context size.

\section{Additional Results} \label{sec:additional_results}

\begin{figure*}
  \vskip 0.2in
  \begin{center}
    \begin{subfigure}[t]{0.45\columnwidth}
      \centering
      \includegraphics[width=\linewidth]{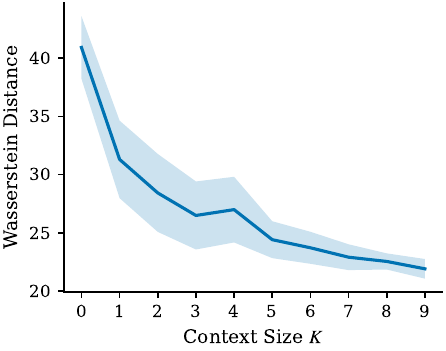}
      \caption{\textbf{Context size scaling.} Wasserstein distance on the melanoma dataset for varying numbers of perturbation experiments in the context set $\mathcal{C}$. Performance improves with the number of experiments $K = |\mathcal{C}|$, where $K=0$ corresponds to conditioning on observational data only. Shaded regions indicate standard deviation over three training seeds.}
      \label{fig:context_sizes}
    \end{subfigure}
    \hspace{0.2in}
    \begin{subfigure}[t]{0.45\columnwidth}
      \centering
      \includegraphics[width=\linewidth]{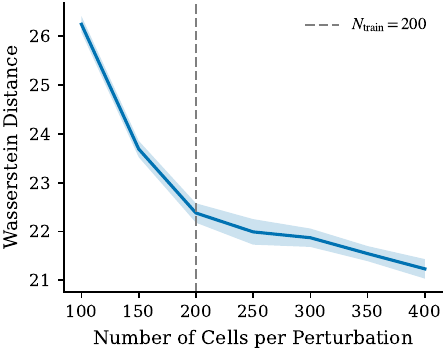}
      \caption{\textbf{Cell set scaling.} Wasserstein distance on the melanoma dataset for varying numbers of cells per perturbation. Performance improves beyond the training configuration (dashed line). A model that does not perform in-context learning would be expected to plateau. Shaded regions indicate standard deviation over ten resampling seeds.}
      \label{fig:cell_scaling}
    \end{subfigure}
    \caption{\textbf{MapPFN scales with more data at inference time.} Both the number of perturbation experiments in context and the number of cells per perturbation improve prediction quality, demonstrating that MapPFN adapts to the available data via in-context learning.}
    \label{fig:inference_scaling}
  \end{center}
  \vskip -0.2in
\end{figure*}

\subsection{Test-time Scaling}

To evaluate how the performance of MapPFN scales with the amount of interventional experiments provided in context, we measure the Wasserstein distance for varying context sizes $K=|\mathcal{C}|$. As shown in \autoref{fig:context_sizes}, test performance improves as additional perturbation experiments are provided in context.

We similarly evaluate how performance scales with the number of cells per perturbation, varying the number of cells at inference time. As shown in \autoref{fig:cell_scaling}, performance improves with more cells, without plateauing at the number of cells seen during training. This suggests that MapPFN can leverage more data by adapting via in-context learning.

\begin{figure*}
  \vskip 0.2in
  \begin{center}
    \centerline{\includegraphics[width=0.9\textwidth]{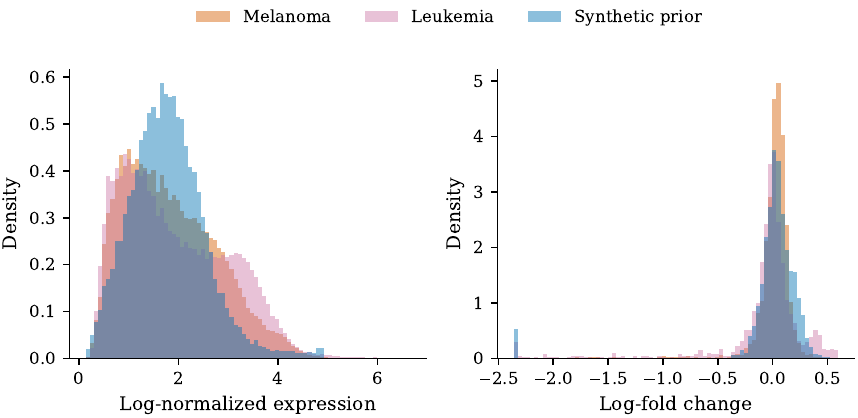}}
    \caption{\textbf{Coverage of the synthetic biological prior over real single-cell data distributions.} Distributions of non-zero expression values (left) and log fold changes (right) for the synthetic biological prior, the melanoma \citep{frangieh_multimodal_2021} and leukemia \citep{papalexi_characterizing_2021} datasets. Log fold change values are clipped to the $[1, 99]$ percentile range for visualization.}
    \label{fig:prior_coverage}
  \end{center}
  \vskip -0.2in
\end{figure*}

\subsection{Prior Coverage}

\autoref{fig:prior_coverage} compares the expression and log fold change distributions of the synthetic biological prior with both real perturbation datasets. Overall, the prior covers the range of expression values and perturbation effects observed in both downstream datasets. For the melanoma dataset \citep{frangieh_multimodal_2021}, the distributional shapes align well, as both the prior and the melanoma gene set consist of genes within a shared regulatory program where all genes are perturbation targets. For the leukemia dataset \citep{papalexi_characterizing_2021}, the expression distribution is bimodal, which we attribute to more than half of the gene set consisting of marker genes that are not themselves perturbation targets. Fine-tuning reduces the distributional gap on both datasets, showing that MapPFN can compensate for distributional differences between synthetic and real data.

\subsection{Architecture Ablation} \label{sec:architecture_ablation}

We compare MMDiT against DiT \citep{peebles_scalable_2023} under the original MapPFN pre-training protocol. We match parameter count by widening DiT at fixed depth, rounded to the nearest width divisible by 8 permitted by the attention kernel. As shown in \autoref{tab:architecture_ablation}, MMDiT performs better on $\text{W}_2$, MR and AUPRC, with MMD, RMSE and PDS falling within standard deviation. These results suggest that per-modality weights are beneficial compared to shared weights.

\begin{table}[htbp]
\caption{\textbf{Architecture ablation comparing MMDiT against DiT.} Performance on the melanoma dataset \citep{frangieh_multimodal_2021}. Mean $\pm$ std over five data splits, each averaged over ten resampling seeds. Bold indicates results within one standard deviation of the best.}
\label{tab:architecture_ablation}
\begin{center}
\resizebox{\textwidth}{!}{%
\begin{tabular}{lccccccc}
\toprule
Architecture & Parameters & $\text{W}_2$ $\downarrow$ & MMD ($\times 10^{-3}$) $\downarrow$ & RMSE $\downarrow$ & PDS $\downarrow$ & MR $\rightarrow 1$ & AUPRC $\uparrow$ \\
\midrule
MMDiT \citep{esser_scaling_2024} & 25.97M & $\textBF{22.25} \pm 1.33$ & $\textBF{63.34} \pm 17.79$ & $\textBF{0.32} \pm 0.04$ & $\textBF{0.22} \pm 0.04$ & $\textBF{1.01} \pm 0.05$ & $\textBF{0.34} \pm 0.07$ \\
DiT \citep{peebles_scalable_2023} & 27.66M & $23.87 \pm 1.58$ & $\textBF{51.47} \pm 18.50$ & $\textBF{0.35} \pm 0.06$ & $\textBF{0.19} \pm 0.04$ & $1.12 \pm 0.04$ & $0.22 \pm 0.08$ \\
\bottomrule
\end{tabular}}
\end{center}
\end{table}

\subsection{Holdout Cell States} \label{sec:holdout_states}

The melanoma dataset comprises three treatment-induced cell states, of which we hold out the IFN-$\gamma$ state in \autoref{tab:singlecell_benchmark}. To verify that our results are not specific to this choice, we repeat the evaluation holding out each cell state. \citet{frangieh_multimodal_2021} report state-specific regulatory programs, with immune-escape programs recovered only in the TIL co-culture and interferon-response programs only under IFN-$\gamma$, so each holdout requires adaptation to a distinct regulatory context. As shown in \autoref{tab:holdout_states}, fine-tuned MapPFN matches or exceeds CondOT and STATE on all three cell states and on all metrics except $\text{W}_2$. Throughout, a single pre-trained checkpoint is adapted both to held-out cell states within a cell line and to unseen cell lines, while every baseline receives supervised training for each holdout.

\begin{table*}[htbp]
\caption{\textbf{Performance across three holdout cell states of the melanoma dataset.} Mean $\pm$ std over five data splits, each averaged over ten resampling seeds. Bold indicates results within one standard deviation of the best for each holdout.}
\label{tab:holdout_states}
\begin{center}
\resizebox{\textwidth}{!}{%
\begin{tabular}{llcccccc}
\toprule
Holdout & Method & $\text{W}_2$ $\downarrow$ & MMD ($\times 10^{-3}$) $\downarrow$ & RMSE $\downarrow$ & PDS $\downarrow$ & MR $\rightarrow 1$ & AUPRC $\uparrow$ \\
\midrule
\multirow[t]{4}{*}{IFN-$\gamma$}
 & CondOT \citep{bunne_supervised_2022} & $22.19 \pm 0.91$ & $\textBF{6.78} \pm 0.33$ & $\textBF{0.10} \pm 0.00$ & $0.06 \pm 0.02$ & $0.04 \pm 0.00$ & $0.41 \pm 0.05$ \\
 & STATE \citep{adduri_predicting_2025} & $\textBF{20.79} \pm 0.83$ & $7.39 \pm 1.19$ & $\textBF{0.09} \pm 0.01$ & $0.17 \pm 0.18$ & $\textBF{0.95} \pm 0.05$ & $0.42 \pm 0.23$ \\
 & MapPFN (pre-trained) & $22.25 \pm 1.33$ & $63.34 \pm 17.79$ & $0.32 \pm 0.04$ & $0.22 \pm 0.04$ & $\textBF{1.01} \pm 0.05$ & $0.34 \pm 0.07$ \\
 & MapPFN (fine-tuned) & $\textBF{21.05} \pm 0.62$ & $\textBF{6.57} \pm 0.56$ & $\textBF{0.09} \pm 0.01$ & $\textBF{0.02} \pm 0.01$ & $\textBF{0.98} \pm 0.01$ & $\textBF{0.62} \pm 0.02$ \\
\midrule
\multirow[t]{4}{*}{TIL co-culture}
 & CondOT \citep{bunne_supervised_2022} & $17.59 \pm 0.41$ & $8.86 \pm 0.66$ & $0.11 \pm 0.01$ & $\textBF{0.06} \pm 0.01$ & $0.06 \pm 0.01$ & $0.49 \pm 0.12$ \\
 & STATE \citep{adduri_predicting_2025} & $\textBF{16.47} \pm 0.36$ & $\textBF{8.28} \pm 0.55$ & $\textBF{0.09} \pm 0.01$ & $0.15 \pm 0.03$ & $0.91 \pm 0.01$ & $0.50 \pm 0.07$ \\
 & MapPFN (pre-trained) & $21.55 \pm 2.25$ & $95.09 \pm 19.08$ & $0.39 \pm 0.05$ & $0.24 \pm 0.03$ & $1.21 \pm 0.11$ & $0.22 \pm 0.11$ \\
 & MapPFN (fine-tuned) & $17.23 \pm 0.37$ & $\textBF{8.59} \pm 0.55$ & $\textBF{0.10} \pm 0.00$ & $\textBF{0.04} \pm 0.02$ & $\textBF{0.98} \pm 0.01$ & $\textBF{0.62} \pm 0.06$ \\
\midrule
\multirow[t]{4}{*}{Control}
 & CondOT \citep{bunne_supervised_2022} & $18.19 \pm 1.22$ & $10.98 \pm 1.66$ & $0.12 \pm 0.02$ & $0.10 \pm 0.02$ & $0.07 \pm 0.02$ & $0.14 \pm 0.03$ \\
 & STATE \citep{adduri_predicting_2025} & $\textBF{16.90} \pm 0.60$ & $\textBF{9.15} \pm 0.79$ & $\textBF{0.08} \pm 0.01$ & $0.11 \pm 0.02$ & $0.91 \pm 0.01$ & $0.29 \pm 0.02$ \\
 & MapPFN (pre-trained) & $20.38 \pm 1.71$ & $58.10 \pm 17.57$ & $0.32 \pm 0.05$ & $0.17 \pm 0.08$ & $1.11 \pm 0.07$ & $0.36 \pm 0.08$ \\
 & MapPFN (fine-tuned) & $\textBF{17.35} \pm 0.50$ & $\textBF{9.47} \pm 1.07$ & $\textBF{0.09} \pm 0.01$ & $\textBF{0.03} \pm 0.01$ & $\textBF{0.96} \pm 0.01$ & $\textBF{0.45} \pm 0.08$ \\
\bottomrule
\end{tabular}}
\end{center}
\end{table*}

\stopcontents[appendix]


\clearpage

\end{document}